\documentclass{article}

\usepackage[preprint]{neurips_2023}
\usepackage{amsthm} 
\usepackage{amsmath}
\usepackage{amssymb} 
\usepackage{thmtools} 
\usepackage{mathtools} 
\usepackage{stmaryrd}

\usepackage{quiver}

\usepackage{xspace}
\usepackage{float}
\usepackage{caption}
\usepackage{subcaption}
\usepackage{enumitem}
\usepackage[framemethod=TikZ]{mdframed}
\usepackage{xcolor}
\usepackage{wrapfig}

\usepackage{algorithm}
\usepackage[noend]{algorithmic}

\newtheorem{definition}{Definition}
\newtheorem{theorem}{Theorem}
\newtheorem{corollary}{Corollary}
\newtheorem{lemma}{Lemma}
\newtheorem{proposition}{Proposition}

\newtheorem{example}{    Example}

\newmdenv[
  linewidth=1pt,
  roundcorner=5pt,
  backgroundcolor=lightgray!20,
  innerleftmargin=10pt,
  innerrightmargin=10pt,
  innertopmargin=10pt,
  innerbottommargin=10pt,
]{greybox}
\newcount\foreachcount

\makeatletter
\let\ea\expandafter
\def\foreachLetter#1#2#3{\foreachcount=#1
  \ea\loop\ea\ea\ea#3\@Alph\foreachcount
  \advance\foreachcount by 1
  \ifnum\foreachcount<#2\repeat}
\def\definecal#1{\ea\gdef\csname c#1\endcsname{\ensuremath{\mathcal{#1}}\xspace}}
\foreachLetter{1}{27}{\definecal}

\makeatletter
\let\ea\expandafter
\def\foreachLetter#1#2#3{\foreachcount=#1
  \ea\loop\ea\ea\ea#3\@Alph\foreachcount
  \advance\foreachcount by 1
  \ifnum\foreachcount<#2\repeat}
\def\definecal#1{\ea\gdef\csname b#1\endcsname{\ensuremath{\mathbf{#1}}\xspace}}
\foreachLetter{1}{27}{\definecal}

\newcommand{\diag}{\mathrm{diag}}

\title{Any Deep ReLU Network is Shallow}

\author{%
  Mattia J. Villani \\
  Department of Informatics\\
  King's College London\\
  London WC2B 4BG \\
  \texttt{mattia.villani@kcl.ac.uk} \\
   \And
  Nandi Schoots \\
  Department of Informatics\\
  King's College London\\
  London WC2B 4BG \\
  \texttt{nandi.schoots@kcl.ac.uk} 
}

\begin{document}

\maketitle

    \begin{abstract}
We constructively prove that every deep ReLU network can be rewritten as a functionally identical three-layer network with weights valued in the extended reals. Based on this proof, we provide an algorithm that, given a deep ReLU network, finds the explicit weights of the corresponding shallow network. The resulting shallow network is transparent and used to generate explanations of the model's behaviour. 
\end{abstract}

\section{Introduction}

Deep learning systems are leveraged for their flexibility and performance to solve real world tasks efficiently; yet, their complexity warrants in-depth analysis to provide safety and soundness guarantees. 
In light of this, we present a method to convert black-box neural networks into functionally identical shallow white-box networks. 

Our contributions are: 
\begin{enumerate}[topsep=0pt]
    \item A constructive proof of the existence of a functionally identical shallow network (three hidden layers) with weights in the extended reals for every deep ReLU network (Section \ref{sec:shallow-network-theorem}); 
    \item An algorithm to find the values of these weights given a trained network (Section \ref{sec:implementation}), which relies on searching the space of possible neuron activations through heuristics that we prove to be exhaustive; and 
    \item Explainability improvements that arise from the shallow network construction (Section \ref{sec:interpretability}) including fast SHAP values.
\end{enumerate}

Our constructive proof permits explicit specification of the weights, thus enabling us to compute the shallow network algorithmically. 
The code for replicating our experiments is provided in the Appendix. 

This work builds on recent breakthroughs that exactly represent ReLU networks as collections of local linear models. 
\citet{sudjianto2020unwrapping} find explicit weights of local linear models for ReLU networks as a function of the weights. Other works compute local approximations of neural networks as explanations \citep{dong2021explain}, or use them to analyse phenomena such as dying neurons 
\citep{Jiang2022}.
Conversely, \citet{He2021} proves that every piece-wise linear function on a bounded domain can be realised as a ReLU network of a given depth, provided as a function of the linear regions.  

 This study extends these findings by enhancing the theoretical understanding of neural networks and establishing a firm connection between linear programming and ReLU activation patterns. Consequently, our work advances global interpretability of neural networks and provides effective heuristics for exploring the extensive search space of activation patterns. 

\section{Background}\label{sec:background}
We restrict our analysis to architectures which are feed-forward neural networks; in particular, those whose activation function for each layer is the rectifier linear unit (ReLU), but generalize our findings in Corollary \ref{cor:mattia-unwrapping} to other ReLU-based architectures. 

\begin{definition}[ReLU Network]
A ReLU network $\mathcal{N} \colon \mathbb{R}^n \rightarrow \mathbb{R}^m$, is a composition of $L \in \mathbb{N}$ hidden layers given by: 
\[ \chi^{(l)} = \sigma( {W^{(l)}} \chi^{(l-1)} + b^{(l)}),\]
where $\sigma$ is an element-wise ReLU activation function, $\sigma(x_i) = \max\{0,x_i\}$. We define $\chi^{(0)} = x$ and the output layer is given by $\mathcal{N}(x) = W^{(L+1)} \chi^{(L)} + b^{(L+1)}$. 
The number of neurons in each layer is given by a vector $\mathbf{N} = [n_1, n_2, \ldots, n_L]$, and all activations are in the positive reals, i.e. $\chi^{(l)} \in \mathbb{R}^{n_l}_{\geq 0}$ for all $l \in \{1,\ldots,L\} = [L]$. We stress the dependence on $x$ by writing $\chi^{(l)}(x)$.
\end{definition}

We can describe these functions as linear models that are applied to certain regions of the input space \citep{sudjianto2020unwrapping}. Moreover, these regions partition $\mathbb{R}^n$. 

\begin{greybox}
\begin{example}
Consider the network given by the weight matrix 
$ W 
= \begin{bmatrix} 1 & 1 \\ 1 & 0  \end{bmatrix}. 
$
Let the input vector be $\begin{bmatrix} x_1 \\ x_2 \end{bmatrix}$. Then, the output of this network, after applying the ReLU activation function, is
$\begin{bmatrix}
y_1 \\
y_2
\end{bmatrix} =
\text{ReLU} \left( \begin{bmatrix}
1 & 1 \\
0 & 1
\end{bmatrix}
\begin{bmatrix}
x_1 \\
x_2
\end{bmatrix} \right) =
\text{ReLU} \left( \begin{bmatrix}
x_1 + x_2 \\
x_2
\end{bmatrix} \right) =
\begin{bmatrix}
\max(0, x_1 + x_2) \\
\max(0, x_2)
\end{bmatrix}.
$

This gives us a partition of the input space $\mathbb{R}^2$. The boundaries are determined by the points where the vector output of the linear transformation has entries equal to zero. 

Then, the four parts (or regions) are given by: \\
\[\omega_1 = \{(x_1,x_2) : x_1 + x_2 > 0, x_2 > 0 \},  
\omega_2 = \{(x_1,x_2) : x_1 + x_2 \leq 0, x_2 > 0 \}, \]
\[ \omega_3 = \{(x_1,x_2) : x_1 + x_2 > 0, x_2 \leq 0 \},
\omega_4 = \{(x_1,x_2) : x_1 + x_2 \leq 0, x_2 \leq 0 \}, \]
such that $\Omega = \{\omega_1, \omega_2,\omega_3,\omega_4\}$ covers the input space.

\begin{figure}[H]
    \begin{center}
    \includegraphics[width=0.28\textwidth]{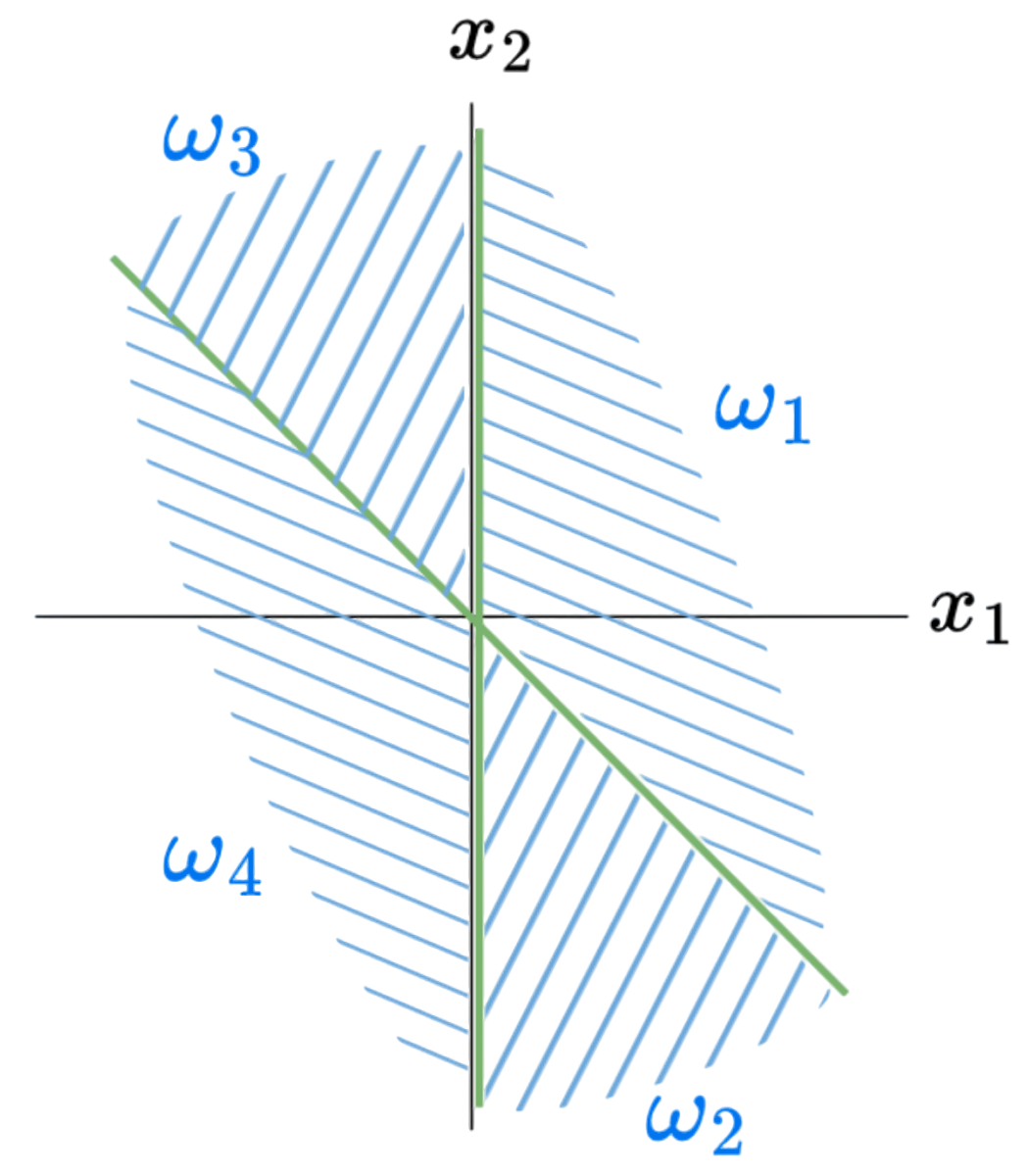}
    \end{center}
\end{figure}
\end{example}
\end{greybox}

\begin{proposition}\label{prop:network-to-partition}[\citet{sudjianto2020unwrapping}]
For a ReLU network $\mathcal{N} \colon \mathbb{R}^n \rightarrow \mathbb{R}^m$ there is a finite partition $\Omega$ of $\mathbb{R}^n$ of cardinality $p := \# \Omega$ such that for each part $\omega \in \Omega$ there exists a piece-wise linear function $f\colon \mathbb{R}^n \rightarrow \mathbb{R}^m$, and its restriction on $\omega$, denoted $f|_\omega$, can be described by a linear function: 
\[f|_\omega (x) = \alpha_\omega^T x + \beta_\omega.\]
Moreover, each part is a  polytope, given by the intersection of a collection of half-spaces.
We write the minimum set of half-space conditions that can be used to specify the entire partition as $H_1,\ldots,H_k$, where each $H_i$ is given by the set of all $x \in \mathbb{R}^n$ such that: 
\[ h_{i,1} x_1 + h_{i,2} x_2 + \ldots + h_{i,n} x_n  > c_i.\]
\end{proposition}

To represent the decomposition and linear models, it is useful to characterise the states of a neural network by looking at which neurons are active and inactive in all layers. 

\begin{definition}[\textbf{Activation Pattern}] For a given ReLU network $\mathcal{N} \colon \mathbb{R}^n \rightarrow \mathbb{R}^m$ with neuron dimensionality vector $\mathbf{N} = [n_1,\ldots,n_L]$, the activation pattern at a point $x \in \mathbb{R}^n$ is a collection of vectors $P = \{P^{(1)},\ldots, P^{(L)}\}$, with $P^{(l)} \in \{0,1\}^{n_l}$ for each $l \in [L]$, such that for all $i \in [n_l]$, 
\[P^{(l)}_i = 1 \iff \chi^{(l)}_i(x) > 0 . \]
\end{definition}

\section{Finding a Shallow Network for a given Piece-wise Linear Function} \label{sec:shallow-network-theorem}

\begin{figure}[h]
    \centering
    \includegraphics[width=0.95\textwidth]{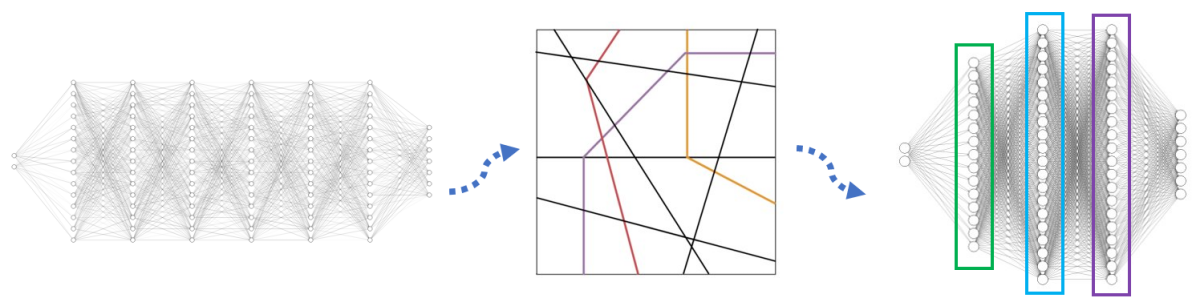}
    \caption{A high-level representation of the conversion of a deep ReLU network (left) using a decomposition of the input space into regions that each have their own linear model (middle) to a shallow ReLU network (right).}
    \label{fig:conversion}
\end{figure}

Figure \ref{fig:conversion} illustrates the intuition that a network can be fully characterized by 1) the partition; and 2) a linear model for each item in the partition. This suggests that we can convert any deep network into a shallow one. We will now formalize and prove this intuition.

\begin{theorem}\label{theorem:main-shallow}
For every ReLU network $\mathcal{N} \colon \mathbb{R}^n \rightarrow \mathbb{R}^m$, there exists a shallow ReLU network $\mathcal{S} \colon \mathbb{R}^n \rightarrow \mathbb{R}^m$ of depth $L = 3$, with weights in the extended reals $\bar{\mathbb{R}} = \mathbb{R} \cup \{\infty\}$ such that for all $x \in \mathbb{R}^n$, the following holds: 
\[\mathcal{N}(x) = \mathcal{S}(x). \]
\end{theorem}

\begin{proof}

Our strategy will be to build a first layer that encodes the linear models and the half-spaces that span the partition. 
We will use the weights that encode the half-spaces to, in the last layer, select the correct linear model out of the list of linear models. 
See Figure \ref{fig:schematic-rep-of-proof} for an overview of the components of the shallow network.
\begin{figure}[h]
    \centering
    \includegraphics[width=0.85\textwidth]{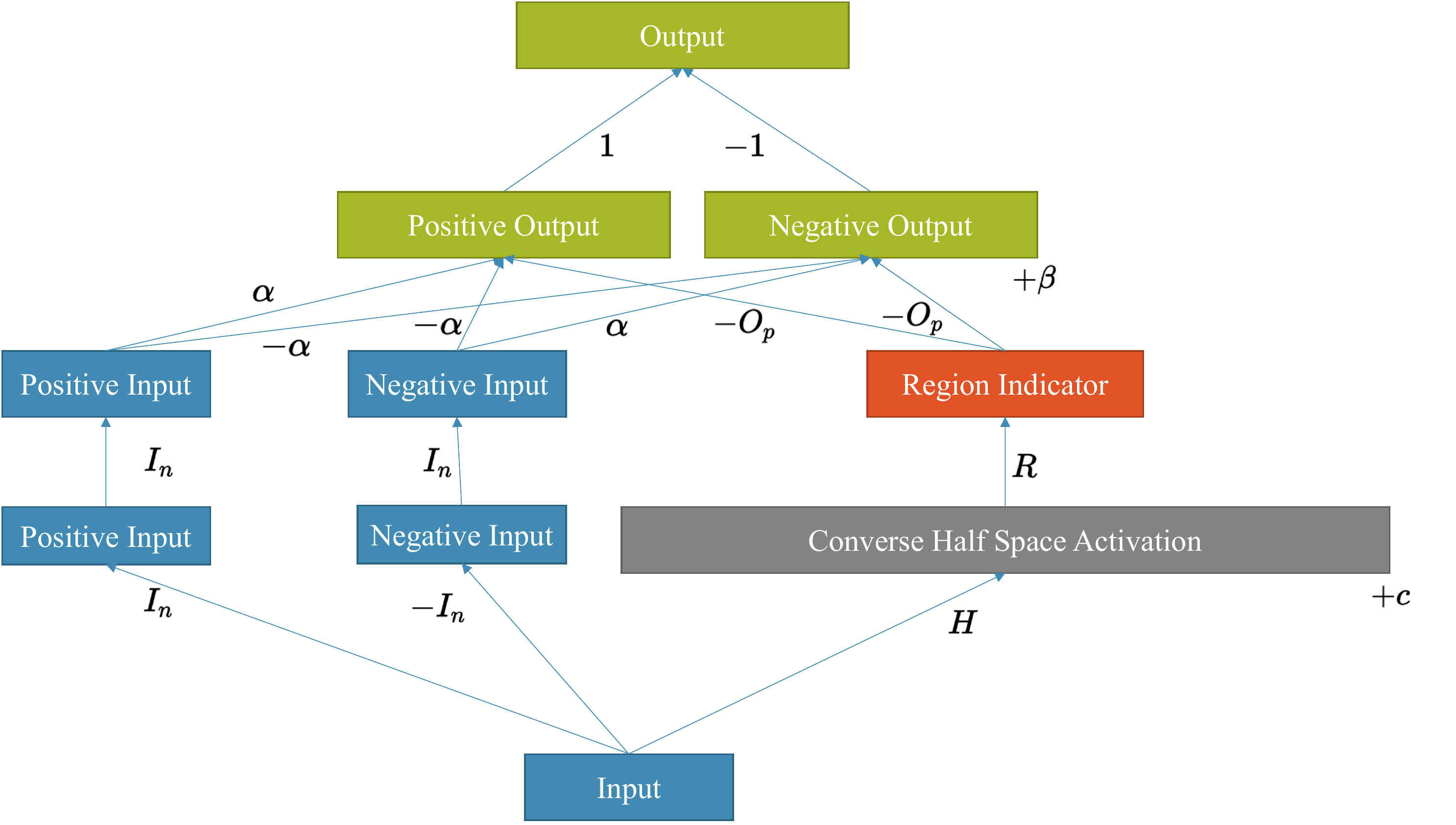}
    \caption{A schematic representation of the shallow network. }
    \label{fig:schematic-rep-of-proof}
\end{figure}

Let $(\{f_\omega\}_{\omega \in \Omega}, \Omega)$ be the decomposition that corresponds to $\mathcal{N}$. Label the parts $1,\ldots,p$. Recall that each part is a polytope and we can represent it as the intersection of half-spaces. In particular, let $H_1,\ldots,H_k$ be the the minimum set of half-spaces that specify the partition, where each $H_i$ is given by the set of all $x \in \mathbb{R}^n$ such that: 
\[ h_{i,1} x_1 + h_{i,2} x_2 + \ldots + h_{i,n} x_n  > c_i.\]
Let $\Psi$ be a linear transformation applying all the half-space conditions: 
\begin{align*}
    \Psi(x) = H x + c 
    = \begin{bmatrix} -h_{1,1} & -h_{1,2} & \ldots & -h_{1,n} \\ -h_{2,1} & -h_{2,2} & \ldots & -h_{2,n}\\ \ldots & & & \ldots \\  - h_{k,1} & - h_{k,2} & \ldots & - h_{k,n} \end{bmatrix} \begin{bmatrix} x_1 \\ x_2 \\\ldots \\ x_n \end{bmatrix}+ \begin{bmatrix} c_1 \\ c_2\\  \ldots \\ c_k \end{bmatrix}, 
\end{align*}
where $H$ is a $k \times n$ matrix. 
Entry-wise this matrix $\Psi(x)$ looks as follows
\[\Psi(x)_{i} = - h_{i,1} \cdot x_1 - h_{i,2} \cdot x_2 -  \ldots - h_{i,n} \cdot x_n + c_{i}.\] 
Notice that this implies that whenever $x\in H_i$ we have $\text{ReLU}(\Psi(x))_{i} = 0$. 
The \textbf{first layer} is given by: 
\[
\chi^{(1)} = \text{ReLU}( W^{(1)}x +b^{(1)})  =\text{ReLU} \left(\begin{bmatrix}
    I_n \\
    - I_n \\
    H 
\end{bmatrix} x + \begin{bmatrix} 0 \\ 0 \\ c \end{bmatrix} \right),
\]
which is an activation vector of length $2\cdot n + k$.

This transformation develops the building blocks to generate neurons that are only inactive when $x$ is in a certain region: we want to construct a layer that only has an inactive neuron whenever the input is in a particular region. 
In other words, we find a $p \times k$ zero-one matrix $R$ which describes the regions in terms of which boundaries are active. Precisely, we can design $R$ so that $R \cdot \text{ReLU}(\Psi(x))$ has a zero entry at $j$ only when $x \in \mathbb{\omega}_j, j \in \{1,\ldots,p\}$. 
Assume without loss of generality that the  half-space conditions corresponding to region $\omega_j$ are $H_1, \ldots , H_a$. The half-space conditions $H_1, \ldots , H_a$ are satisfied at a point $x$ if and only if $x \in \omega_j$.
We construct $R$ such that at row $j$ it assigns a 1 to entries $1, \ldots, a$, and a 0 to all other entries in that row, repeating this for all regions.
This leads to the description of the \textbf{second layer}: 
\begin{align*}
\chi^{(2)} 
= \text{ReLU}( W^{(2)} \chi^{(1)} + b^{(2)}) 
&= \text{ReLU} 
\left(\begin{bmatrix} 
I_{n} & 0 & 0 \\ 
 0& I_{n} & 0 \\
 0& 0 & R \end{bmatrix} \chi^{(1)} + \begin{bmatrix}
    0 \\ \vdots  \\ 0
\end{bmatrix} \right) \\
& = \text{ReLU} \left(\begin{bmatrix} \hspace{12pt} \text{ReLU}( x )  \\
\hspace{12pt} \text{ReLU}( - x) \\
R \cdot \text{ReLU}(\Psi(x)) \end{bmatrix} \right), 
\end{align*} 
as an activation vector of length $2 \cdot n + p$.
 
Finally, in the third layer we 1) multiply the input with each of the linear models, and 2) use the indicator function that is zero only at entry $p$ if the input $x$ is in part $p$, to isolate the correct linear model. Without loss of generality, we assume that $m = 1$ so that $\mathcal{N}: \mathbb{R}^n \rightarrow \mathbb{R}$. If the dimension of the output space $m >1$, then we repeat the construction of the last layer $m$ times. 

Let $\Lambda$ be a linear transformation given by: 
\[
\Lambda(x) = \alpha^T x + \beta,
\]
where $\alpha, \beta$ are respectively a $p \times n$ matrix and a $p$-dimensional bias, such that the row $\alpha_i$ and entry $\beta_i$ contain the coefficients of $f_i$. Note that, in this step, if $m > 1$, then the matrix and vector would be $p \cdot m \times n$ and $p \cdot m$ dimensional respectively.

We construct a matrix $ O_{p} = \theta \cdot I_{p}$, where $ \theta = \infty$, such that $\infty \cdot 0 = 0$ and $\infty \cdot a = \infty$ for all $a \in \mathbb{R}_{>0}$.
Note that Python contains such a constant, denoted by \texttt{inf}. 
We use this matrix so that when we subtract it from a matrix of linear models, the only unaffected linear model will be $f_j$, i.e.
\[\Lambda(x) - O_p \cdot \text{ReLU} (R \cdot \text{ReLU}(\Psi(x))) 
= \begin{bmatrix} - \infty \\ \vdots \\ -\infty \\\Lambda(x)_j \\ - \infty \\ \vdots \\ - \infty \end{bmatrix},\] 
which is a vector of length $p$ for which the only positive entry is given by the linear model at the region $j$ evaluated at $x$. 
Since elements of other regions can lie arbitrarily close to region $\omega_j$, each $R\cdot \text{ReLU} (\Psi(x))$ can be arbitrarily small, and so we cannot just choose a very large number instead of infinity. 
This entails our \textbf{third layer} activations are given by:   
\begin{align*}
\chi^{(3)} 
= 
\text{ReLU}(W^{(3)} \chi^{(2)} + b^{(3)}) 
& = \text{ReLU} \left(\begin{bmatrix} \alpha & -\alpha  & - O_{p} \\ -\alpha 
& \alpha  & - O_{p}  \end{bmatrix} \chi^{(2)} + \begin{bmatrix} \beta \\ -\beta \end{bmatrix}\right) \\
& = \text{ReLU} \left( \begin{bmatrix} \alpha & -\alpha  & - O_{p} \\ -\alpha & \alpha  & - O_{p}  \end{bmatrix} \begin{bmatrix} \hspace{12pt} \text{ReLU}(x) \\ \hspace{12pt} \text{ReLU}( - x) \\ R \cdot \text{ReLU}(\Psi(x)) \end{bmatrix} + \begin{bmatrix} \beta \\ -\beta \end{bmatrix}\right) \\
& = \text{ReLU} \left( \begin{bmatrix} \hspace{3.5pt} \Lambda(x)  \hspace{4pt} - O_{p} \cdot \text{ReLU} \left(R \cdot \text{ReLU}(\Psi(x)) \right) \\ -\Lambda(x) - O_{p} \cdot \text{ReLU}\left(R \cdot \text{ReLU}(\Psi(x))\right) \end{bmatrix} \right)
\end{align*}
which is a vector of length $2 p$ with exactly one non-zero entry at either $j$ or $2j$. This non-zero entry is $\Lambda(x)$ at row $j$ (if $\Lambda(x)$ is positive) or $-\Lambda(x)$ at row $2j$ (if $\Lambda(x)$ is negative). 

Lastly we need a weight matrix to project onto $\mathbb{R}^m$. This projection requires $2p$ weights. When we project the $-\Lambda(x)$ value to the output layer, we need to swap the sign. The first $p$ weights of $W^{(4)}$ are 1 and the last $p$ weights are $-1$: 
\[ \mathcal{S}(x) = W^{(4)} \cdot \chi^{(3)}  = \Lambda(x)_j = \mathcal{N}(x).\]

Note that here we assumed $m=1$. In general, the length of $\chi^{(3)}$ is $2 \cdot p \cdot m$, as is the number of required projection weights.
\end{proof}

\begin{corollary}
Given a ReLU network $\mathcal{N} \colon \mathbb{R}^n \rightarrow \mathbb{R}^m$, the shallow network $\mathcal{S} \colon \mathbb{R}^n \rightarrow \mathbb{R}^m$ of depth $L = 3$, 
as specified in the proof of Theorem \ref{theorem:main-shallow}, has width bounded by \[  \max \{ 2n+k, 2n+p, 2\cdot p \cdot m \}, \]
where $p = \# \Omega$ and $k = \# \{ \text{half space conditions} \}$.
\end{corollary}
Note that \citet{montufar2014number},  \citet{Serra2020}, and \citet{Chen2022}, find progressively tighter bounds for the number of linear regions in ReLU networks as functions of the number of neurons at each layer.
\begin{corollary}
For every continuous piece-wise linear function $f$, there exists a shallow neural network $\mathcal{S}$ of depth $L = 3$ with weights in the extended reals $\bar{\mathbb{R}} = \mathbb{R} \cup \{\infty\}$ such that for all $x \in \mathbb{R}^n$ a compact subspace of $\mathbb{R}^n$, the functions are identical.
\end{corollary}

\begin{proof}
\cite{He2021} proves that every piece-wise linear function on a compact subspace of real space can be represented exactly by a neural network of finite depth and width.
\end{proof}

\begin{corollary}\label{cor:mattia-unwrapping}
    For every tensor convolutional neural network (including all graph convolutional neural networks, recurrent neural networks and convolutional neural networks) with ReLU activation functions there exists a shallow neural network $\mathcal{S}$ of depth $L= 3$ with weights in the extended reals $\bar{\mathbb{R}} = \mathbb{R} \cup \{\infty\}$ such that for all $x \in \mathbb{R}^n$, the functions are identical. 
\end{corollary}

\begin{proof}
\citet{villani2023unwrapping} show that tensor convolutional networks with ReLU activations can be decomposed into local linear models (like feedforward ReLU networks).
\end{proof}

This theory allows us to verify whether two networks are functionally identical on their domain of definition, rather than just on their test dataset.
By lexicographically orderering the linear models and half-space conditions, the algorithm maps any set of functionally identical networks to the same unique shallow network.

\section{Implementation} \label{sec:implementation}

Given a deep ReLU network, we propose an algorithm that finds the weights of a functionally identical shallow network. 
This algorithm is based on the constructive proof of Theorem \ref{theorem:main-shallow}. We show that every activation pattern leads to a linear program, whose feasible set is exactly a region of the partition of a ReLU network. 

In order to find the weights of the shallow network, we need to search the space of activation patterns to find all the activation patterns that yield a nonempty region. This is a vast space, with $2^{n_1 + \ldots+ n_L}$ possible patterns, which we reduce through certain heuristics. 
We develop a theory that connects linear programs to ReLU networks, and search the reduced space of patterns via a brute force algorithm.  

\subsection{Formal Correspondence between Linear Programs and Regions}

We define local linear programs as the set of constraints required by an activation pattern at a given layer.

\begin{definition}[Local linear program]
Let $P = \{ P^{(1)}, \ldots , P^{(L)} \}$ be an activation pattern.
The \textit{local linear program}  corresponding to $P^{(l)}$, 
is given by the set of activation vectors $\chi^{(l-1)}$ at layer $(l-1)$ such that
\[ \left(\diag(\mathbf{1} - P^{(l)})\cdot W^{(l)} - \diag(P^{(l)})\cdot W^{(l)}  \right) \cdot \chi^{(l-1)} 
\leq  - \diag(\mathbf{1}-P^{(l)})\cdot b^{(l)} + \diag(P^{(l)})\cdot b^{(l)}.\]
\end{definition}

The following lemma is used in the proof of Lemma \ref{lemma:empty-feass-set} and Proposition \ref{prop:global-lin-prog}.

\begin{lemma}
Let $x \in \mathbb{R}^n$, $x$ has activation pattern $P^{(l)}$ at layer $l$ only if $\chi^{(l-1)}$ satisfies the local linear program corresponding to $P^{(l)}$. \textbf{(See Appendix.)} 
\end{lemma}

We use the following lemma to reduce the search space of possible patterns.

\begin{lemma}\label{lemma:empty-feass-set}
Let $P = \{P^{(1)},\ldots, P^{(l)}, \ldots, P^{(L)} \}$ be the activation pattern of $\omega_P$. If the corresponding local linear program at layer $l$ has an empty feasible set, then $\omega_P = \emptyset$. \textbf{(See Appendix.)}
\end{lemma}

We define global linear programs in terms of the input vector as opposed to in terms of an activation vector. 
\begin{definition}[Global linear program]
Let $P = \{ P^{(1)}, \ldots , P^{(L)} \}$ be an activation pattern.
Recursively define
\begin{align*}
  A^{(1)} & := W^{(1)}  \\
  A^{(l)} & := W^{(l)} \cdot \diag (P^{(l-1)}) \cdot A^{(l-1)} \\
  d^{(1)} & :=  b^{(1)} \\
  d^{(l)} & :=  \cdot W^{(l)} \cdot \diag(P^{(l-1)}) \cdot d^{(l-1)}  + b^{(l)}.
\end{align*}
The \textit{global linear program} corresponding to $P$ at layer $l$, 
is given by the set of inputs $x\in \mathbb{R}^n$ such that for all $1\leq l \leq L$ we have 
\[ \left(\diag(\mathbf{1} - P^{(l)})\cdot A^{(l)} - \diag(P^{(l)})\cdot A^{(l)}  \right) \cdot x 
\leq  - \diag(\mathbf{1}-P^{(l)})\cdot d^{(l)} + \diag(P^{(l)})\cdot d^{(l)}.\]
\end{definition}

The following proposition is used in the algorithm to find the parameters of the hyperplanes as the constraints of the global linear program.

\begin{proposition}\label{prop:global-lin-prog}
If $x \in \mathbb{R}^n$ has activation pattern $P = \{ P^{(1)}, \ldots , P^{(L)} \}$,
then $x$ satisfies the global linear program corresponding to $P$. \textbf{(See Appendix.)}
\end{proposition}

\subsection{Algorithm}\label{sec:algorithm}
It is sufficient to find $H,c, \alpha, \beta$ to build the shallow network, while knowing which half spaces are needed to define each region. The algorithm we present to find the weights is structured as follows. First, we identify the set of feasible patterns. For each of these, we compute the parameters $A^{(l)}, d^{(l)} \forall l \in [L]$ of the global linear program. Stacking these matrices determines the half-space conditions for each activation pattern, and hence $-H = \texttt{stack}(\{A^{(l)}\}_{l \in [L]})$ and similarly for $c = \texttt{stack}(\{d^{(l)}\}_{l \in [L]})$. Finally, given the model weights and an activation pattern, it is easy to compute the local linear model $\alpha, \beta$ by using the explicit formula from \cite{sudjianto2020unwrapping}.   
\begin{algorithm*}
\caption{Find Shallow Network Weights}
\begin{algorithmic}[1] 
\REQUIRE $L > 0, \forall l \in [L], n_i > 0, W^{(l)} \in \mathbb{R}^{n_l\times n_{l+1}}, b^{(l)} \in \mathbb{R}^{n_{l+1}} $ \COMMENT{this is the input}
\ENSURE $\alpha, \beta, H,c $ \COMMENT{this is the output}
\STATE{$\forall l \text{ initialise lists called }\texttt{layerwise-feasible-patterns}_l$}
\FOR{$l \in [L]$}
\FOR{$P^{(l)} \in \{0,1\}^{n_l}$}
\IF{ $\texttt{has-nonempty-local-linear-program} (P^{(l)}, W^{(l)}, b^{(l)}) = \text{True}$}
\STATE{\text{store} $P^{(l)}$ in $\texttt{layerwise-feasible-patterns}_l$}
\ENDIF
\ENDFOR
\ENDFOR
\STATE{\text{Initialise lists of lists called \texttt{feasible-patterns}, \texttt{all-hyperplanes}}}
\STATE{$\text{store } \texttt{layerwise-feasible-patterns}_1 \text{ in } \texttt{ feasible-patterns } $}
\FOR{$l \in \{2,...,L\}$}
\FOR{$P^{(l)} \text{ in } \texttt{layerwise-feasible-patterns}_l$}
\FOR{$\texttt{pattern} \text{ in } \texttt{feasible-patterns}$}
\STATE{$A^{(l)}, d^{(l)} \leftarrow \texttt{get-linear-program-weights}(P^{(l)}, \{W^{(i)}, b^{(i)}\}_{i\in[l]})$}
\IF{$\texttt{has-nonempty-global-linear-program}(P^{(l)}, \{A^{(i)}, d^{(i)}\}_{i \in [l]}) = \text{True}$}
\STATE{$\text{store } P^{(l)} \text{ in } \texttt{pattern}$}
\STATE{$\text{store } \texttt{pattern} \text{ in } \texttt{feasible-patterns}$}
\STATE{$\text{store } A^{(l)}, d^{(l)} \text{ in } \texttt{pattern}$}
\STATE{$H,c \leftarrow \texttt{get-hyperplanes}(\{A^{(i)}, d^{(i)}\}_{i \in [l]})$}
\STATE{$\text{store } H, c \text{ in } \texttt{hyperplanes}$}
\ENDIF
\ENDFOR
\ENDFOR
\ENDFOR
\FOR{$P \leftarrow \texttt{pattern} \text{ in } \texttt{feasible-patterns}$}
\STATE{$\alpha_P, \beta_P \leftarrow \texttt{get-local-linear-model}(P,\{W^{(l)},b^{(l)})\}_{l \in [L]}) $}
\STATE{store $\alpha_P, \beta_P$ in \texttt{local-linear-models}}
\ENDFOR
\RETURN \texttt{local-linear-models}, \texttt{hyperplanes}
\end{algorithmic}
\label{alg:algorithm}
\end{algorithm*}

While the algorithm has complexity $\mathcal{O}(2^{\Sigma_{l=1}^L n_l})$, \citet{Hanin2019} find that the number of activation patterns is generally small. 
Our algorithm avoids searching the entire pattern space by pruning large portions of it.

\begin{wrapfigure}{r}{0.5\textwidth}
    \vspace{+0cm}
    \includegraphics[width=0.5\textwidth]{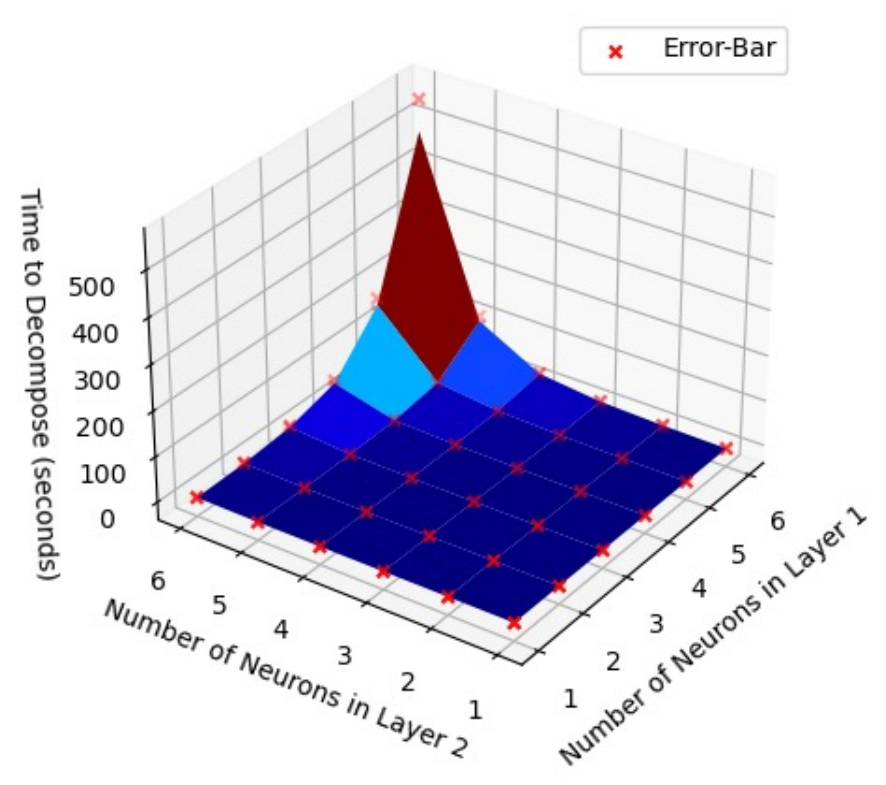}
    \caption{We plot the time to compute the shallow networks corresponding to randomly initialized networks $\mathcal{N}$ on the z-axis as a function of the number of neurons in the first hidden layer (y-axis) and the second hidden layer (x-axis) of $\mathcal{N}$. Error bars indicate one standard deviation.}
    \label{fig:running-time}
    \vspace{+0cm}
\end{wrapfigure}

We executed the algorithm on a local machine's CPU
(16 GB RAM). 
We randomly initialized weights on a small network (using Xavier uniform). 
The networks have three hidden layers, with variable first and second layer width and a fixed third layer width.
We measure the time it takes to compute a shallow network for each of the randomly initialized networks.
For each combination of layer widths we repeat the experiment five times. 

The time taken to compute the algorithm for randomly initialized networks grows exponentially with the number of neurons, see Figure \ref{fig:running-time}. 
Note that the initialisation forces the algorithm to work in the worst case, where none of the activation patterns can be pruned and all patterns need to be checked.
As expected based on the algorithm's design, the time grows faster in the number of second layer nodes than in first layer nodes. In Figure \ref{fig:running-time}, we see that 6 neurons in the first layer, but only 5 neurons in the second layer requires slightly less than 100 seconds to compute, whereas the reverse takes slightly more than 100 seconds on average.


\section{Interpretability}\label{sec:interpretability}

The finding that any ReLU Network can be decomposed into a set of local linear models \citep{sudjianto2020unwrapping} has made black-box neural networks more interpretable. 
Very recently, an algorithm was introduced \cite{balestriero2023fast} that (like our algorithm) provides a global interpretation of the model by specifying the partition and the linear model of each part.

The fully understood construction of the white-box shallow network empowers users to perform model surgery on it.
Below, we highlight two ways in which interpretability is enhanced: 
(1) we can compute fast SHAP values for each datapoint; 
(2) given a region, we give an explanation for the region by approximating it with a hypercube containing it.

\subsection{Calculating Exact SHAP Values}

SHAP values are popular metrics used as explanations in the field of explainable artificial intelligence \citep{lundberg2017unified, Samek2021}.
For a given datapoint they impute how much each input dimension contributes to the output. SHAP values are defined as follows. 

\begin{definition}[SHAP values]
\label{def:shap}
    Given a function $f\colon \mathbb{R}^n \rightarrow \mathbb{R}^m$ and an input $x\in \mathbb{R}^n$.
    A function $f_x \colon \{0,1\}^n \rightarrow \mathbb{R}^m$ is a \emph{masking function} if it evaluates $f(x)$ with $x$ unchanged when $z_i = 1$ and evaluates a `turned off' input  at dimension $i$ (such as mean or zero) when the input $z_i = 0$, for all $i \in [n]$. 
    The \emph{SHAP values} for each feature (or dimension) $i \in [n]$ are given by an  $n\times m$ matrix.
    Let $f_x \colon \{0,1\}^n \rightarrow \mathbb{R}^m$ be a masking function.
    The SHAP values are then: 
    \[\phi_i(f, x) = \sum_{z \in \{z' \in \{0,1\}^n \colon z'_i =1\} }  \frac{|z|! (n-|z| -1)!}{n!}[f_x(z) - f_x(z_{-i} )],\]
    where 
    $(z_{-i})_j := z_j$ for all $j$ excluding $i$; and $|z|$ counts the number of nonzero entries of $z$. 
\end{definition}

Typically, for machine learning models, these SHAP values are estimated. 
However, for a linear model they are known \citep{lundberg2017unified} and are given in terms of the mean of a distribution over the input space.
Hence, given a decomposition of a network we can give exact SHAP values.

\begin{proposition}
Let $\mathcal{N} \colon \mathbb{R}^n \rightarrow \mathbb{R}^m$ be a ReLU network with decomposition into local linear models. Let $\omega$ be a region with local linear model $f(x') = \alpha x' + \beta$, let $D$ be a distribution over $\omega$ and let $x \in \omega$. Let $f_x$ be a masking function, where given a masked dimension 
we define $f_x$ as returning a sample from the distribution $D$. Then the SHAP values for feature $i$ are given by 
\[ \phi_i(\mathcal{N}, x) = \alpha ( x - \mathbb{E}_{D(\omega)}(x)).\]
\end{proposition}

\begin{proof}
Since $\mathcal{N}$ is linear on the polytope $\omega$, for every sampled input $X \sim D(\omega)$ we will have that  $\mathcal{N}(X) = f_x(X) = \alpha X + \beta$. The result then follows from  \cite{lundberg2017unified}. 
\end{proof}

A consequence of this proposition is that SHAP values can be computed exactly and almost instantly from the decomposition after computing the mean of the background distribution. \footnote{This can be done analytically if we choose the distribution to be uniform. However, it may be inappropriate to select this distribution, since the polytope may not be bounded.}


\subsection{Explainability via Hypercubes around Regions}

Given a region, we can approximate it by drawing the smallest possible hypercube around it. This enables us to summarise the region, using a small number of parameters: the location and size of the hypercube. 
In particular, we highlight that this summary is in tension with faithfully representing the region, which would require specifying many hyperplanes and $n$ parameters for each one of these hyperplanes. 

\begin{figure}[ht]
    \vspace{-0.3cm}
    \centering
    \includegraphics[width=1.0\textwidth]{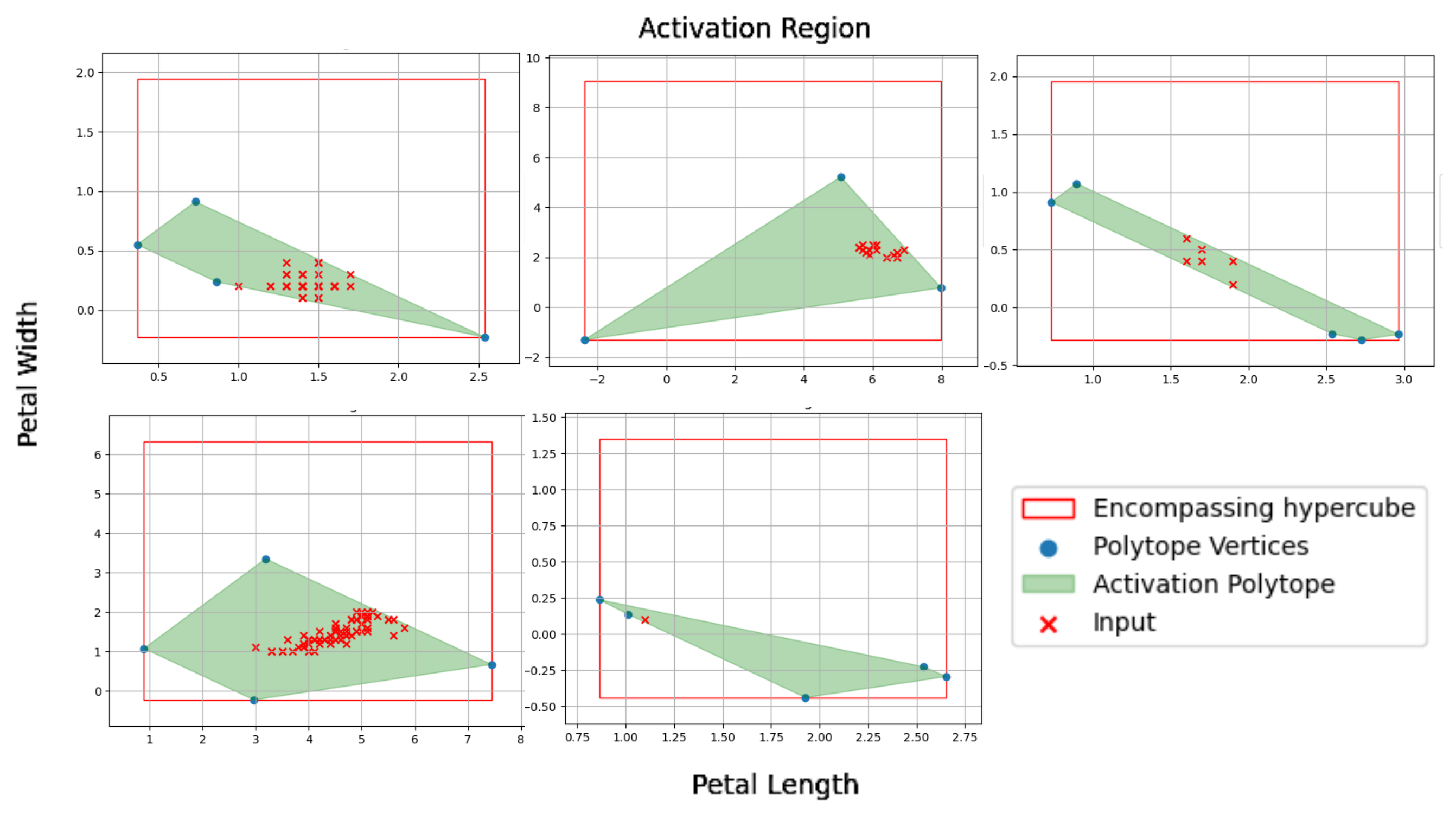}
    \vspace{-0.5cm}
    \caption{We plot the regions (in green) for a test portion (40\%) of the first two dimensions of the Iris dataset. 
    All these datapoints (red crosses) fall in one of five regions. In red we draw the smallest hypercube that fits the entire region. 
    The x-axis represents petal length, and the y-axis petal width. }
    \label{fig:plot-regions}
    \vspace{-0.2cm}
\end{figure}

We trained a feedforward ReLU network with neuron vector [5,7,4] on the first two dimensions of the Iris dataset \citep{iris-dataset} for 5000 epochs with a Sigmoid output layer, cross entropy loss and Adam with learning rate 0.0001 and betas equal to 0.9 and  0.999.
We then used our algorithm, as introduced in Section \ref{sec:algorithm} to decompose the neural network and find all regions. We then select the regions where test data lives and record them in Figure \ref{fig:plot-regions}. 
The hypercubes are coarse explanations that become less faithful as the number of dimensions increases. 
However, identifying the position 
as well as inspecting which inputs are in a region helps us identify common characteristics of the inputs; 
the size parameter of the cube can be compared to others in order to capture an approximation of the size of the region.



\section{Discussion}



Our contribution has been to prove all deep ReLU networks can be rewritten as a shallow network (Theorem \ref{theorem:main-shallow}), while providing a way to compute the weights algorithmically (Algorithm \ref{alg:algorithm}).
This enhances interpretability by allowing us to quickly compute feature importance and giving  details of the exact, local functional relationship between input and output.
Deep neural networks discern hierarchical concepts, with later layers encapsulating more complex concepts than earlier ones. 
Remarkably, we demonstrate that these concepts can be rearranged horizontally, maintaining their original functionality. 
This restructuring allows us to derive a shallow network from a deep one.

Our construction of a shallow network is not injective, in that multiple (functionally identical) deep networks with different architectures can all map to the same shallow network. 
If we consider equivalence classes of neural networks that each encode a different function, then our shallow networks can be seen as a representative of their equivalence class. 



\subsection{Limitations}

The construction of a shallow network makes use of infinite weights.
Although these weights can be implemented in \texttt{Python}, they can not be fine-tuned using gradient descent. 
Note that the weights encoding the linear models can be fine-tuned. 
To maintain the transparent network structure after fine-tuning, the weights encoding identity matrices between the blue blocks in Figure \ref{fig:schematic-rep-of-proof} should be frozen. 
Additionally, the four weight matrices denoted by $\alpha$ and $-\alpha$ between the blue and green blocks should be updated in conjunction with one another.  

The current algorithm is well suited for small models, but large computational resources are needed to perform the search of feasible patterns in larger models. Future work should focus on finding efficient heuristics to prune the space of possible activation patterns. Moreover, research should prioritise developing faithful explanations that navigate the complexity of these models.

    \bibliographystyle{agsm}
	\bibliography{references.bib}

    \newpage

    \section*{Appendix}
\setcounter{section}{7} 
\setcounter{subsection}{0}
\renewcommand{\thesubsection}{\thesection.\arabic{subsection}} 



\subsection{Proofs supporting Section \ref{sec:implementation} } \label{app:proofs}

\begin{definition}[Local linear program]
Let $P = \{ P^{(1)}, \ldots , P^{(L)} \}$ be an activation pattern.
The \textit{local linear program}  corresponding to $P^{(l)}$, 
is given by the set of activation vectors $\chi^{(l-1)}$ at layer $(l-1)$ such that
\[ \left(\diag(\mathbf{1} - P^{(l)})\cdot W^{(l)} - \diag(P^{(l)})\cdot W^{(l)}  \right) \cdot \chi^{(l-1)} 
\leq  - \diag(\mathbf{1}-P^{(l)})\cdot b^{(l)} + \diag(P^{(l)})\cdot b^{(l)}.\]
\end{definition}

\begin{lemma}\label{lemma:loc-lin-prog-n}
Let $x \in \mathbb{R}^n$; $x$ has activation pattern $P^{(l)}$ at layer $l$ only if $\chi^{(l-1)}$ satisfies the local linear program corresponding to $P^{(l)}$.  
\end{lemma}

\begin{proof}
Suppose $x \in \mathbb{R}^n$ has activation pattern $P^{(l)}$.\\
If $P^{(l)}_i = 0$
then $( W^{(l)} \chi^{(l-1)}(x)  + b^{(l)} )_i \leq 0$
then $( W^{(l)} \chi^{(l-1)}(x) )_i \leq - b^{(l)}_i$, 
so 
\[\diag(\mathbf{1} - P^{(l)})\cdot W^{(l)}  \cdot \chi^{(l-1)}(x)  
\leq  - \diag(\mathbf{1}-P^{(l)})\cdot b^{(l)} .\]
If $P^{(l)}_i = 1$
then $( W^{(l)} \chi^{(l-1)}(x) + b^{(l)} )_i \geq 0$
then $ - (W^{(l)} \chi^{(l-1)} (x))_i \leq b^{(l)}_i$,
so
\[- \diag(P^{(l)})\cdot W^{(l)} \cdot \chi^{(l-1)}(x)  \leq \diag(P^{(l)})\cdot b^{(l)}.\]
We can conclude that
\[ \left(\diag(\mathbf{1} - P^{(l)})\cdot W^{(l)} - \diag(P^{(l)})\cdot W^{(l)}  \right) \cdot \chi^{(l-1)}(x) 
\leq  - \diag(\mathbf{1}-P^{(l)})\cdot b^{(l)} + \diag(P^{(l)})\cdot b^{(l)}.\]
\end{proof}

\begin{lemma}\label{lemma:empty-feas-set}
Let $P = \{P^{(1)},..., P^{(l)}, ..., P^{(L)} \}$ be the activation pattern of $\omega_P$. If the corresponding local linear program at layer $l$ has an empty feasible set, then $\omega_P = \emptyset$. 
\end{lemma}

\begin{proof}
We will prove this lemma by contradiction.
Assume that the local linear program at layer $l$ has an empty feasible set and that $w_P \neq \emptyset$.

Let $x \in \omega_P$.
Then $x$ has activation pattern $P^{(l)}$.
By Lemma \ref{lemma:loc-lin-prog-n} this means that 
$x$ satisfies
\[ \left(\diag(\mathbf{1} - P^{(l)})\cdot W^{(l)} - \diag(P^{(l)})\cdot W^{(l)}  \right) \cdot \chi^{(l-1)}(x) 
\leq  - \diag(\mathbf{1}-P^{(l)})\cdot b^{(l)} + \diag(P^{(l)})\cdot b^{(l)}.\]
However, in this case the local linear program at layer $l$ has non-empty set, so we have found a contradiction.
\end{proof}

\begin{definition}[Global linear program]
Let $P = \{ P^{(1)}, \ldots , P^{(L)} \}$ be an activation pattern.
Recursively define
\begin{align*}
  A^{(1)} & := W^{(1)}  \\
  A^{(l)} & := W^{(l)} \cdot \diag (P^{(l-1)}) \cdot A^{(l-1)} \\
  d^{(1)} & :=  b^{(1)} \\
  d^{(l)} & :=  \cdot W^{(l)} \cdot \diag(P^{(l-1)}) \cdot d^{(l-1)}  + b^{(l)}.
\end{align*}
The \textit{global linear program} corresponding to $P$ at layer $l$, 
is given by the set of inputs $x\in \mathbb{R}^n$ such that for all $1\leq l \leq L$ we have 
\[ \left(\diag(\mathbf{1} - P^{(l)})\cdot A^{(l)} - \diag(P^{(l)})\cdot A^{(l)}  \right) \cdot x 
\leq  - \diag(\mathbf{1}-P^{(l)})\cdot d^{(l)} + \diag(P^{(l)})\cdot d^{(l)}.\]
\end{definition}

\begin{proposition}

If $x \in \mathbb{R}^n$ has activation pattern $P = \{ P^{(1)}, \ldots , P^{(L)} \}$,
then $x$ satisfies the global linear program corresponding to $P$.
\end{proposition}

\begin{proof}
By definition of $P^{(l)}$ we have that,  $\chi^{(l)} = \diag(P^{(l)})\cdot (W^{(l)} \chi^{(l-1)} + b^{(l)})$ for $x \in \omega_P$.
We will prove by induction that
\[\chi^{(l)} = \diag(P^{(l)}) \left(A^{(l)} x + d^{(l)}\right),    \forall x \in \omega_P.\]
The base case is true by definition: 
\[ \chi^{(1)} = \diag(P^{(1)}) \cdot \left( A^{(1)} x + d^{(1)} \right) = \diag(P^{(1)}) \cdot \left( W^{(1)} x + b^{(1)} \right).\]
Now assuming that $\chi^{(l-1)} = \diag(P^{(l-1)} )\left( A^{(l-1)} x + d^{(l-1)} \right) $ we will show that $\chi^{(l)} = \diag(P^{(l)}) \left(A^{(l)} x + d^{(l)}\right)$.
We write
\begin{align*}
\chi^{(l)} 
&= \diag( P^{(l)}) \left(W^{(l)} \chi^{(l-1)} + b^{(l)} \right)\\
&= \diag( P^{(l)}) \left( W^{(l)} \left(A^{(l-1)} x + d^{(l-1)} \right) + b^{(l)}\right) \\
&= \diag( P^{(l)}) \left( W^{(l)} A^{(l-1)} x + W^{(l)} d^{(l-1)} + b^{(l)}\right) \\
&= \diag(P^{(l)}) \left( A^{(l)} x + d^{(l)} \right).
\end{align*}

By Lemma \ref{lemma:loc-lin-prog-n}, for all $l$, $\chi^{(l-1)}$ satisfy the local linear program, or equivalently all $\diag(P^{(l-1)}) \left( A^{(l-1)}x + d^{(l-1)}\right)$ satisfy the linear program if $x \in \omega_P$. We rewrite the local linear programs as: 

\begin{align*} \left(\diag(\mathbf{1} - P^{(l)})\cdot W^{(l)} - \diag(P^{(l)})\cdot W^{(l)}  \right) \diag(P^{(l-1)}) \left( A^{(l-1)}x + d^{(l-1)}\right) \\
 \leq  - \diag(\mathbf{1}-P^{(l)})\cdot b^{(l)} + \diag(P^{(l)})\cdot b^{(l)}.
\end{align*}
From which we conclude that, 
\begin{align*}
& \left(\diag(\mathbf{1} - P^{(l)})\cdot W^{(l)} - \diag(P^{(l)})\cdot W^{(l)}  \right) \diag(P^{(l-1)}) A^{(l-1)}x  \\
&\leq  - \diag(\mathbf{1}-P^{(l)})\cdot (b^{(l)} + W^{(l-1)}\diag(P^{(l-1}) d^{(l-1)}) + \diag(P^{(l)})\cdot (b^{(l)}+ W^{(l-1)} \diag(P^{(l-1)}) d^{(l-1)}), 
\end{align*}
simplifying to:
\begin{align*}
& \left(\diag(\mathbf{1} - P^{(l)})\cdot A^{(l)} - \diag(P^{(l)})\cdot A^{(l)}  \right) x  \leq  - \diag(\mathbf{1}-P^{(l)})\cdot d^{(l)} + \diag(P^{(l)})\cdot d^{(l)}, 
\end{align*}
as required. 
\end{proof}

\end{document}